\documentclass[runningheads]{llncs}
\usepackage[T1]{fontenc}
\usepackage{graphicx,verbatim}
\usepackage{booktabs}
\usepackage{multirow}
\usepackage{amssymb}
\usepackage{array}
\usepackage{amsmath}
\usepackage{cite}
\usepackage{pifont}
\usepackage[table]{xcolor}
\usepackage{tabularx}
\newcolumntype{Y}{>{\centering\arraybackslash}X}
\usepackage{hyperref}
\usepackage{color}

\hypersetup{
    colorlinks=true,
    citecolor=blue,
    linkcolor=blue,
    urlcolor=blue
}
\begin{document}
\title{CGSM: Concept-Guided Segmentation Model for Precise Pulmonary Lesion Delineation}
\titlerunning{Concept-Guided Segmentation Model for Lesion Delineation}
%

\author{
Changheng Lin\inst{\dagger}
\and
Wenjie Zhang\inst{\dagger}
\and
Yushan Lu
\and
Xinyue Yan
\and
Xiao Jia\inst{*}
\and
Wei Zhang\inst{*}
}

\authorrunning{C. Lin et al.}

\institute{
Shandong University, China
}

\maketitle              
\begin{abstract}
Accurate segmentation of pulmonary lesions is essential for effective clinical diagnosis and treatment strategies. Existing segmentation approaches often lack task-specific semantic guidance, as text-based annotations typically offer coarse localization of lesions, leading to inadequate delineation of lesion boundaries and poor performance on small-scale lesions. To address this, we propose CGSM, a Concept-Guided Segmentation Model that integrates LLM-generated and clinically reviewed concepts into the segmentation process. Specifically, we design a Concept-Visual Alignment Module (CVAM) to activate relevant tokens within the concepts that align with visual features, enhancing the interaction between textual and visual information. In addition, we introduce a Concept Modulated Decoder (CM-Decoder), which uses concepts from CVAM as modulation signals to facilitate the adaptive fusion of image and text features, improving the segmentation accuracy. Extensive experiments on two public datasets show that CGSM achieves state-of-the-art performance, with results of 91.59\% Dice and 84.49\% mIoU on the QaTa-COV19 dataset, demonstrating its effectiveness in pulmonary lesion segmentation.


\end{abstract}
\section{Introduction}

Thoracic imaging, particularly chest X-rays (CXR) and computed tomography (CT), is essential for the diagnosis and monitoring of pulmonary diseases~\cite{wang2020temporal, chung2020ct}. Despite advances in deep learning–based radiological analysis \cite{greenspan2016guest, fan2024deep}, accurate lesion segmentation remains challenging due to heterogeneous lesion morphology, subtle contrast, and complex anatomical backgrounds. As a result, models trained on images alone~\cite{ronneberger2015u, zhou2018unet++, isensee2021nnu, cao2022swin, wang2022uctransnet} may have limited ability to capture radiology-informed priors, such as lesion shape, margin characteristics, texture, and spatial distribution.

To better incorporate such radiology-informed priors, recent studies have explored multi-modal learning that integrates accompanying textual information~\cite{huang2021gloria, kim2021vilt, yang2022lavt, tomar2022tganet, hu2024lga}. Textual guidance provides complementary semantic cues and has been shown to improve segmentation performance. For example, Li et al.~\cite{li2023lvit} proposed LViT, which leverages localized text annotations and a hybrid architecture of CNN–Transformer for vision-language fusion, outperforming conventional mono-modal segmentation methods. Zhong et al.~\cite{zhong2023ariadne} proposed GuideDecoder, which performs hierarchical cross-modal fusion throughout the decoder and improves segmentation precision.

Despite these advances, the textual supervision available in current pipelines is often coarse and typically focuses on location-oriented descriptions, lacking task-relevant attributes closely tied to boundary delineation, such as margin sharpness, internal texture, and distribution patterns. Consequently, the alignment between lesion appearance and textual cues is often weak or ambiguous, limiting segmentation accuracy for lesions with heterogeneous morphology and subtle contrast. Moreover, fine-grained alignment between text and image regions is challenging because textual descriptions typically lack explicit spatial grounding, which may introduce ambiguity when linking phrases to image regions.

To address these limitations, we propose a Concept-Guided Segmentation Model (CGSM) that enriches coarse clinical text with radiology-informed concepts generated by a large language model (LLM). These concepts introduce boundary-relevant morphological and textural priors, providing explicit guidance for lesion delineation. Our main contributions are summarized as follows:
\begin{itemize}
    \item[-] We introduce CGSM, a concept-guided segmentation framework that enriches coarse clinical text with radiology-informed priors generated by an LLM.
    \item[-] We design a Concept-Visual Alignment Module (CVAM) to learn structured visual–concept correspondences, enforcing semantic consistency between visual tokens and clinical concepts.
    \item[-] We design a Concept Modulated Decoder (CM-Decoder) to inject aligned concepts into cross-modal decoding via adaptive modulation, enabling morphologically grounded boundary delineation.
    \item[-] Experiments on two public datasets demonstrate that CGSM achieves state-of-the-art performance and improves segmentation accuracy across multiple evaluation metrics. 
\end{itemize}

\section{Method}
\begin{figure}[!t] 
    \centering
    \includegraphics[width = \textwidth]{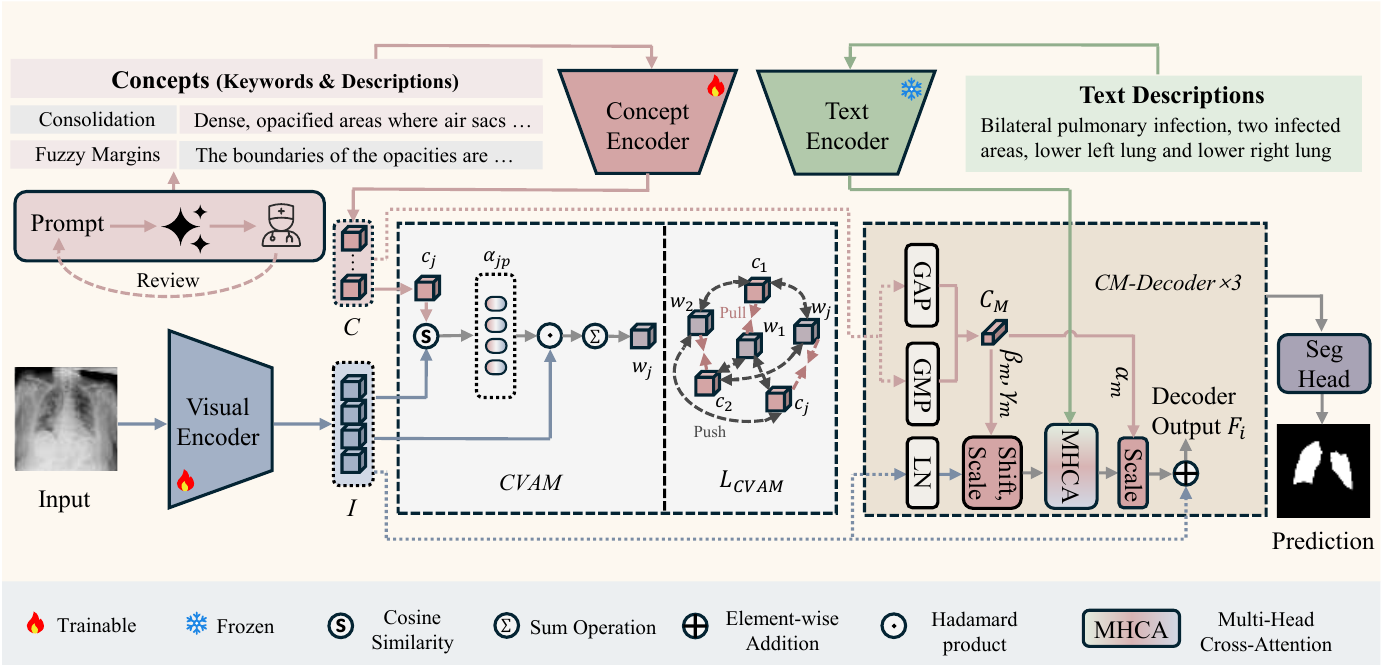} 
    \caption{\textbf{Overview of CGSM:} Our CGSM consists of three independent encoders and a Concept Modulated Decoder (CM-Decoder). The Concept-Visual Alignment Module (CVAM) enforces the consistency between each concept token and supporting visual evidence through a loss function, and subsequently, the aligned concepts guide the cross-modal fusion of image and text features in the CM-Decoder hierarchically. }
    \label{fig1}
\end{figure}

As illustrated in Fig.~\ref{fig1}, we propose a Concept-Guided Segmentation Model (CGSM) for medical image segmentation. CGSM consists of a visual encoder, a text encoder, a concept encoder, and a Concept Modulated Decoder (CM-Decoder). Given an image-text pair, the visual encoder extracts multi-scale visual features $I$, while the text and concept encoders produce textual embeddings $T$ and concept tokens $C$, respectively. A Concept–Visual Alignment Module (CVAM) aligns each concept token with its supporting visual evidence via contrastive learning, yielding image-grounded concepts. These aligned concepts are then injected into the CM-Decoder as modulation signals to guide hierarchical cross-modal fusion, enabling accurate lesion segmentation.

\subsection{LLM-Assisted Concept Construction}
\begin{figure}[!t] 
    \centering
    \includegraphics[width = \textwidth]{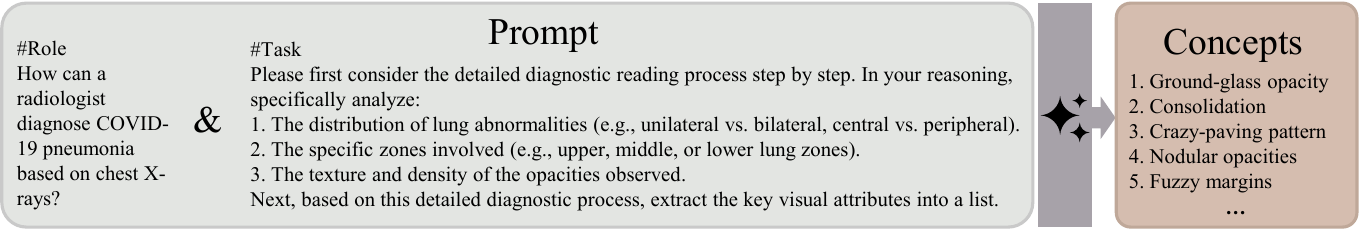} 
    \caption{Example of the prompt used for LLM-based concept generation and the resulting concept phrases.}
    \label{fig2}
\end{figure}

We construct a curated concept set as structured clinical priors through an LLM-assisted, clinician-verified pipeline. For a target disease within a specific imaging modality, an LLM (Gemini-3~\cite{team2023gemini}) is prompted to generate a concise list of concept phrases that capture key lesion attributes, including morphology, edge sharpness, internal texture, and spatial distribution. An example of the LLM prompt and generated concept phrases is shown in Fig.~\ref{fig2}.
The candidate concepts are then independently rated by five clinicians on a 10-point scale. Concepts with an average score below 6 are discarded and re-generated with refined prompts until the quality criterion is satisfied. This process yields a high-quality concept set jointly developed by the LLM and clinicians, which is subsequently used as structured concept tokens in our model.

\subsection{Encoder Design}
Given an input image $X \in \mathbb{R}^{H \times W \times 3}$, we adopt ConvNeXt-Tiny~\cite{liu2022convnet} as the visual encoder to extract a four-level feature pyramid $\{I_i\}_{i=1}^{4}$, where $I_i\in\mathbb{R}^{H_i\times W_i\times d_i}$. 
We employ a pretrained CXR-BERT~\cite{boecking2022making} for language encoding. For the paired text, we directly use a fully frozen CXR-BERT to extract contextual token features $T\in\mathbb{R}^{L\times 768}$, where $L$ denotes the token length. For the concept phrases, we reuse the same frozen CXR-BERT as the backbone and further attach a trainable concept projection head $\phi$ to adapt concept representations to the segmentation task. Formally, $C=\phi(\mathrm{CXR\text{-}BERT}(P))$, where $P$ denotes the tokenized concept phrases and $C=\{c_j\}_{j=1}^{N_t}$ with each $c_j\in\mathbb{R}^{768}$, and $N_t$ is the number of concept tokens.

\subsection{Concept-Visual Alignment Module (CVAM)}
To ground the concept tokens in visual features and highlight task-relevant concepts during decoding, we propose a Concept-Visual Alignment Module (CVAM) that explicitly aligns concept tokens with patch-level image features. Given the last-stage patch features of the visual encoder $ \{v_p\}_{p=1}^{N_v} $ and the concept token embeddings $C=\{c_j\}_{j=1}^{N_t}$, CVAM establishes token-to-patch correspondences via similarity-based attention. Our goal is to identify the most relevant visual regions for each clinical concept. Specifically, we first compute the cosine similarity between each concept token and all visual patches, and normalize over patches to obtain attention weights:
\begin{align}
s_{jp} &= \mathrm{sim}(c_j, v_p), \\
\alpha_{jp} &= \mathrm{softmax} \!\left( \frac{s_{jp}}{\tau} \right),
\end{align}
where $\tau$ is a temperature parameter. The resulting concept-conditioned visual evidence $w_j$ is then synthesized as a weighted sum of all patches: 
\begin{equation}
w_j=\sum_{p=1}^{N_v}\alpha_{jp}\,v_p.
\end{equation}
While this soft-attention grounding mechanism captures localized evidence, it is susceptible to feature collapse, where different concept tokens attend to similar redundant visual regions. To enforce token-discriminative representations, we introduce a bidirectional contrastive loss $L_{\text{CVAM}}$ based on the InfoNCE formulation~\cite{oord2018representation}. This objective pulls matched concept-visual pairs $(w_j, c_j)$ together while pushing apart mismatched pairs simultaneously:
\begin{equation}
L_{\text{CVAM}} = \sum_{j=1}^{N_t} \left( \ell_{w \to c}^{(j)} + \ell_{c \to w}^{(j)} \right),
\end{equation}
where $\ell_{w \to c}^{(j)}$ and $\ell_{c \to w}^{(j)}$ can be represented as:
\begin{equation}
\ell_{w \to c}^{(j)} = -\log \frac{\exp(\text{sim}(w_j, c_j) / \tau)}{\sum_{k=1}^{N_t} \exp(\text{sim}(w_j, c_k) / \tau)}
\end{equation}
\begin{equation}
\ell_{c \to w}^{(j)} = -\log \frac{\exp(\text{sim}(w_j, c_j) / \tau)}{\sum_{k=1}^{N_t} \exp(\text{sim}(w_k, c_j) / \tau)}
\end{equation}
By minimizing $L_{\text{CVAM}}$, the model encourages each visual evidence $w_j$ to align with its corresponding clinical concept $c_j$. This alignment provides image-grounded concept representations for the CM-Decoder, facilitating more accurate and interpretable segmentation.

\subsection{Concept Modulated Decoder (CM-Decoder)}
Building upon the aligned concept representations, we design a Concept Modulated Decoder (CM-Decoder) that integrates visual features $I_i$, textual features $T$, and concept features $C$ through concept-conditioned normalization and cross-modal interaction.

At each decoding stage, text and concept embeddings are first linearly projected to match the channel dimension of the visual features $I_i$. To obtain a compact global descriptor from the set of concept tokens, we employ global pooling operations over the channel-aligned concept features $\hat{C}$. Specifically, we adopt Global Average Pooling (GAP) and Global Max Pooling (GMP):
\begin{equation}
C_{\mathrm{GAP}} = \mathrm{GAP}\!\left(\mathrm{GELU}(\hat{C} W_1)\right), \quad
C_{\mathrm{GMP}} = \mathrm{GMP}\!\left(\mathrm{GELU}(\hat{C} W_2)\right),
\end{equation}
where $W_1$ and $W_2$ are learnable projection matrices. While $C_{\mathrm{GAP}}$ captures global contextual statistics, $C_{\mathrm{GMP}}$ emphasizes the most salient concept activations. The two are further fused to form a unified global concept descriptor:
\begin{equation}
C_M = \mathrm{MLP}\!\left([C_{\mathrm{GAP}}, C_{\mathrm{GMP}}, C_{\mathrm{GAP}} + C_{\mathrm{GMP}}]\right),
\end{equation}
where $[\cdot \, , \, \cdot , \, \cdot]$ denotes concatenation. This aggregation captures both contextual statistics and salient concept activations.

Instead of using fixed learnable scaling and bias parameters in Layer Normalization, we introduce concept-conditioned modulation driven by the concept representation $C_M$. Specifically, three modulation parameters are regressed from $C_M$ via a regression network based on linear layers, whose output is subsequently split into a channel-wise scaling factor $\gamma_m$, a bias term $\beta_m$, and a residual scaling coefficient $\alpha_m$. These parameters dynamically control how visual features are normalized and fused with the text context. For the $i$-th decoder layer, the visual feature $I_i$ is first normalized and adaptively modulated as:
\begin{equation}
I_{\mathrm{mod}}^i 
= \mathrm{LN}(I_i) \odot (1 + \gamma_m) + \beta_m,
\end{equation}
where $\mathrm{LN}(\cdot)$ denotes Layer Normalization and $\odot$ represents the Hadamard product. Unlike standard normalization, where scaling and bias parameters are fixed after training, the proposed modulation allows them to vary conditioned on $C_M$, thereby injecting clinically relevant priors into the feature space.

The modulated feature is subsequently integrated with textual representations via multi-head cross-attention:
\begin{equation}
F_i = I_i + \alpha_m \, \mathrm{MHCA}(I_{\mathrm{mod}}^i, T),
\end{equation}
where $\mathrm{MHCA}(\cdot,\cdot)$ denotes the multi-head cross-attention operation, $T$ represents textual embeddings, and $F_i$ is the output of the $i$-th decoder layer. The residual coefficient $\alpha_m$ further regulates the contribution of cross-modal interaction, enabling adaptive control over the strength of concept guidance.

Through concept-conditioned normalization and adaptive residual scaling, the decoder dynamically balances visual evidence and semantic priors, leading to more robust and semantically consistent segmentation.

\paragraph{\textbf{Overall Objective Function:}} 
The proposed CGSM is optimized using a composite objective function that jointly enforces pixel-wise accuracy, region-level consistency, and concept–visual alignment. The overall loss $L_{total}$ is formulated as:
\begin{equation}
L_{total} = \lambda_{1}L_\mathrm{CE} + \lambda_{2}L_\mathrm{Dice} + \lambda_{3} L_\mathrm{CVAM} \,,
\end{equation}
where $\lambda_{1}$, $\lambda_{2}$, and $\lambda_{3}$ are weighting coefficients that balance the contribution of each term. $L_{\mathrm{CE}}$ and $L_{\mathrm{Dice}}$ denote the Cross-Entropy loss and the Dice loss, which supervise the classification of the pixel-level and the consistency of the region-overlap, respectively. $L_{\mathrm{CVAM}}$ is introduced to enhance semantic alignment between concept representations and visual features, thereby promoting concept-aware segmentation.

\section{Experiments and Results}
\subsection{Datasets}
We evaluate our method on two public datasets, QaTa-COV19~\cite{degerli2022osegnet} and MosMedData+~\cite{morozov2020mosmeddata}. QaTa-COV19 contains $9,258$ COVID-19 CXRs with expert pixel-level lesion annotations and predefined train/val/test splits ($5,716$/$1,429$/$2,113$). MosMedData+ consists of $2,729$ CT slices ($2,183$/$273$/$273$), each with ground-truth lesion masks. Following Li et al.~\cite{li2023lvit}, both datasets provide paired structured medical reports, enabling multi-modal segmentation.
\subsection{Implementation Details}
Input images are resized to $224 \times 224$. Data augmentation includes random zooming with a probability of $0.1$. The network is optimized using the AdamW optimizer with a batch size of $32$. A cosine annealing schedule is adopted to decay the learning rate from $3\times10^{-4}$ to $1\times10^{-6}$. The weighting coefficients in the overall objective are empirically set to $\lambda_{1} = 1.0$, $\lambda_{2} = 1.0$, and $\lambda_{3} = 0.5$ unless otherwise specified. The visual backbone is initialized with ImageNet~\cite{deng2009imagenet}-pretrained weights. The text encoder is kept frozen to preserve stable semantic representations. All experiments are implemented in PyTorch~\cite{paszke2019pytorch} with PyTorch Lightning and MONAI~\cite{cardoso2022monai}, and conducted on an NVIDIA A100 SXM4 40GB GPU.
\begin{table}[!t]
    \centering
    \caption{Quantitative comparison of different methods on QaTa-COV19 and MosMedData+ datasets. Best and second-best results are bolded and underlined.}
    \label{tab:performance_comparison}
    
    \renewcommand{\arraystretch}{1}
    
    \setlength{\tabcolsep}{6pt} 
    \begin{tabularx}{\textwidth}{l c YYYY}
        \toprule
        \multirow{2}{*}{\textbf{Method}} & \multirow{2}{*}{\textbf{Venue}} & \multicolumn{2}{c}{\textbf{QaTa-COV19}} & \multicolumn{2}{c}{\textbf{MosMedData+}} \\
        \cmidrule(lr){3-4} \cmidrule(lr){5-6}
        & & \textbf{Dice} $\uparrow$ & \textbf{mIoU} $\uparrow$ & \textbf{Dice} $\uparrow$ & \textbf{mIoU} $\uparrow$ \\
        \midrule
        
        U-Net\cite{ronneberger2015u} & MICCAI'15 & 79.02 & 69.46 & 64.60 & 50.73 \\
        U-Net++\cite{zhou2018unet++} & MICCAI'18 & 79.62 & 70.25 & 71.75 & 58.39 \\
        nnUNet\cite{isensee2021nnu}  & Nature'21 & 80.42 & 70.81 & 72.59 & 60.36 \\
        
        
        GLoRIA\cite{huang2021gloria} & ICCV'21 & 79.94 & 70.68 & 72.42 & 60.18 \\
        TGANet\cite{tomar2022tganet}  & MICCAI'22    & 79.87 & 70.75 & 71.81 & 59.28 \\
        LViT\cite{li2023lvit}        & IEEE TMI'23 & 83.66 & 75.11 & 74.57 & 61.33 \\
        LGA\cite{hu2024lga}          & MICCAI'24 & 84.65 & 76.23 & 75.63 & 62.52 \\
        RecLMIS\cite{huang2024cross} & IEEE TMI'24 & 84.84 & 76.44 & 77.48 & \underline{65.07} \\
        GuideDecoder\cite{zhong2023ariadne} & MICCAI'23 & 89.78 & 81.45 & 77.75 & 63.60 \\
        TGCAM\cite{guo2024common}    & MICCAI'24 & 90.60 & 82.81 & 77.82 & 63.69 \\
        MMI-UNet\cite{bui2024visual} & MICCAI'24 & \underline{90.88} & \underline{83.28} & \underline{78.42} & 64.50 \\
        
        \rowcolor{gray!20}
        \textbf{CGSM (Ours)} &  & \textbf{91.59} & \textbf{84.49} & \textbf{79.08} & \textbf{65.40} \\
        \bottomrule
    \end{tabularx}
\end{table}

\begin{figure}[!t] 
    \centering
    \includegraphics[width = \textwidth]{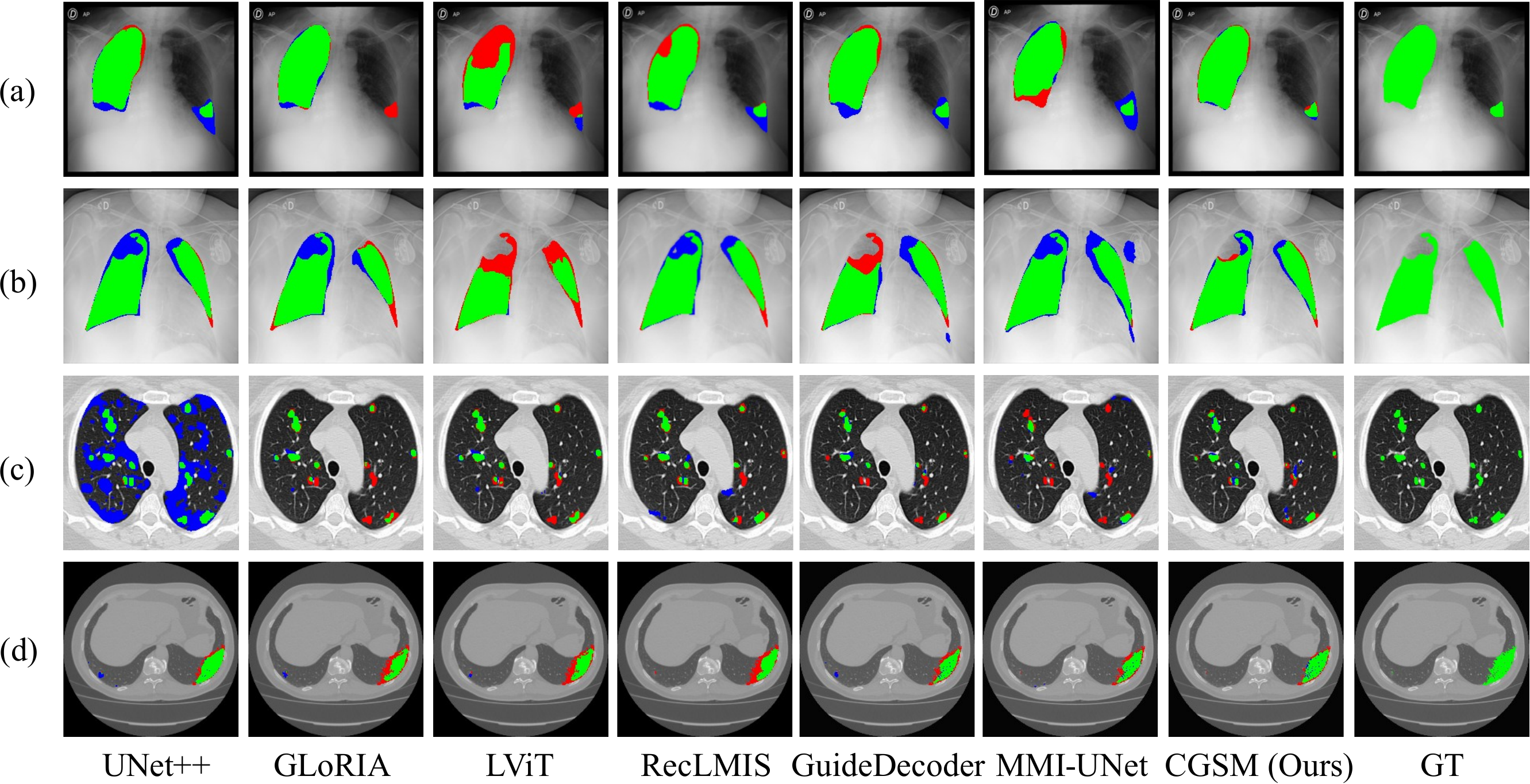} 
    \caption{Qualitative results of segmentation models on two datasets, QaTa-COV19 (a–b) and MosMedData+ (c–d). Green, red, and blue indicate true positives, false negatives, and false positives.} 
    \label{fig3}
    
\end{figure}

\subsection{Comparison Experiments}
As presented in Table~\ref{tab:performance_comparison}, we compare our proposed CGSM with state-of-the-art medical image segmentation methods on the QaTa-COV19 and MosMedData+ datasets. Our method consistently outperforms existing approaches across all metrics on both datasets. Specifically, compared to the current state-of-the-art method, MMI-UNet~\cite{bui2024visual}, CGSM improves the Dice score by 0.71\% on QaTa-COV19 and 0.66\% on MosMedData+, and the mIoU by 1.21\% and 0.90\%, respectively. Qualitative evaluation (Fig. \ref{fig3}) demonstrates that CGSM excels in precise boundary delineation and small-lesion detection. This efficacy is attributed to the integration of concept-guided features and adaptive modulation, which effectively incorporate clinical priors to refine feature representations.

\begin{table}[!t]
    \centering
    \caption{Ablation studies on the QaTa-COV19 and MosMedData+ datasets.}
    \label{tab:ablation_study}
    
    \renewcommand{\arraystretch}{1}
    
    \setlength{\tabcolsep}{3pt} 
    
    \begin{tabularx}{\textwidth}{l cc YYYY}
        \toprule
        \multirow{2}{*}{\textbf{Variant}} & \multirow{2}{*}{\textbf{CVAM}} & \multirow{2}{*}{\textbf{CM-Decoder}} & \multicolumn{2}{c}{\textbf{QaTa-COV19}} & \multicolumn{2}{c}{\textbf{MosMedData+}} \\
        \cmidrule(lr){4-5} \cmidrule(lr){6-7}
        & & & \textbf{Dice} $\uparrow$ & \textbf{mIoU} $\uparrow$ & \textbf{Dice} $\uparrow$ & \textbf{mIoU} $\uparrow$ \\
        \midrule
        
        (a) Baseline      & \ding{55} & \ding{55} & 90.32 & 82.53 & 77.91 & 63.65 \\
        (b) +Concept & \ding{55} & \ding{55} & 90.89 & 83.34 & 78.39 & 64.55 \\
        (c) +CVAM         & \ding{51} & \ding{55} & 91.16 & 83.78 & 78.63 & 64.85 \\
        (d) +CM-Decoder   & \ding{55} & \ding{51} & 91.25 & 83.91 & 78.81 & 64.97 \\
        
        \rowcolor{gray!20}
        (e) \textbf{CGSM (Full)} & \ding{51} & \ding{51} & \textbf{91.59} & \textbf{84.49} & \textbf{79.08} & \textbf{65.40} \\
        \bottomrule
    \end{tabularx}
\end{table}

\subsection{Ablation Study}
Table \ref{tab:ablation_study} presents the results of the ablation study on the QaTa-COV19 and MosMedData+ datasets. Compared to the baseline model (variant a), introducing concept tokens (variant b) yields a slight performance improvement. 
Although concept tokens improve semantic understanding and segmentation accuracy, they alone are insufficient to fully leverage contextual information. The addition of CVAM (variant c) further boosts performance, demonstrating that aligning visual features with concept tokens effectively bridges the gap between image content and conceptual knowledge, leading to improvements in segmentation accuracy. Incorporating the CM-Decoder (variant d) results in an additional increase in both Dice and mIoU metrics. The CM-Decoder facilitates the integration of visual and conceptual information via adaptive modulation, improving segmentation performance by better incorporating prior knowledge and context into the feature representations. The full CGSM (variant e), which combines CVAM and CM-Decoder, achieves the highest performance across all variants. This demonstrates the synergistic effect of the two components, leading to the most significant improvements in both Dice and mIoU across both datasets.

\section{Conclusion}
In this paper, we present CGSM, a concept-guided multi-modal framework for medical image segmentation. By introducing radiology-informed concepts, CVAM aligns concept tokens with visual evidence, while the CM-Decoder injects aligned concepts to guide cross-modal fusion. Experiments on two public datasets demonstrate consistent improvements over existing methods. These results suggest that concept-level priors provide an effective and lightweight mechanism for multi-modal medical segmentation.

%
%
%
\bibliographystyle{splncs04}
\bibliography{main}
%




\end{document}